\documentclass[fleqn,10pt]{wlscirep}
\usepackage[utf8]{inputenc}
\usepackage[T1]{fontenc}
\usepackage{subcaption}

\title{TS-SatFire: A Multi-Task Satellite Image Time-Series Dataset for Wildfire Detection and Prediction}

\author[1]{Yu Zhao}
\author[1]{Sebastian Gerard}
\author[1, *]{Yifang Ban}
\affil[1]{KTH Royal Institute of Technology, Stockholm, 11428, Sweden}
\affil[*]{Corresponding Author: yifang@kth.se}

\begin{abstract}
Wildfire monitoring and prediction are essential for understanding wildfire behaviour. With extensive Earth observation data, these tasks can be integrated and enhanced through multi-task deep learning models. We present a comprehensive multi-temporal remote sensing dataset for active fire detection, daily wildfire monitoring, and next-day wildfire prediction. Covering wildfire events in the contiguous U.S. from January 2017 to October 2021, the dataset includes 3552 surface reflectance images and auxiliary data such as weather, topography, land cover, and fuel information, totalling 71 GB. Each wildfire’s lifecycle is documented, with labels for active fires (AF) and burned areas (BA), supported by manual quality assurance of AF and BA test labels. The dataset supports three tasks: a) active fire detection, b) daily burned area mapping, and c) wildfire progression prediction. Detection tasks use pixel-wise classification of multi-spectral, multi-temporal images, while prediction tasks integrate satellite and auxiliary data to model fire dynamics. This dataset and its benchmarks provide a foundation for advancing wildfire research using deep learning.
\end{abstract}
\begin{document}

\flushbottom
\maketitle
%
%
\thispagestyle{empty}

\section*{Background $\&$ Summary}

Wildfires and the associated loss of forests have shown an increasing trend in recent years \cite{Andela2017AHD, Curtis2018ClassifyingDO, Tyukavina2022GlobalTO}. Monitoring and understanding the behaviour of wildfires are essential for mitigating this natural hazard. Satellite-based solutions for wildfire monitoring have been widely investigated in the remote sensing community, including active fire detection and burned area mapping. Active fire detection involves identifying the location of active fires from satellite imagery, while burned area mapping entails detecting the total burned area from satellite imagery. Wildfire progression prediction aims to forecast the progression of the wildfire based on its current location and environmental information, such as weather, fuel, and topography.

Current satellite-based wildfire products concentrate on active fire detection and burned area mapping \cite{Wooster2021SatelliteRS, Giglio2009AnAB, viirs, modis, col6, LizundiaLoiola2020ASA}. However, the multi-criteria thresholding method employed by active fire products often results in unreliable detections. Also, the accuracy and the temporal resolution of the current burned area mapping product are limited. Various efforts have been made to improve detection accuracy using deep learning \cite{de2021active, Ban2020NearRW}. These studies suggest that deep learning models can substantially improve detection accuracy by leveraging the spectral, spatial, and temporal information present in satellite datacubes. Accurate detection of active fires and burned areas provides crucial current-status information about wildfires, serving as the foundation for wildfire prediction tasks. In recent years, several works have investigated wildfire progression prediction using remote sensing data \cite{NEURIPS2023_ebd54517, 9840400}. These works aim to use auxiliary data along with remote sensing images to predict wildfire progression. However, most of the dataset targets a single task although the same sensor is used. There are limited works that simultaneously cover detection and prediction tasks.
\begin{figure}[!hbt]
    \centering
    \includegraphics[width=\linewidth]{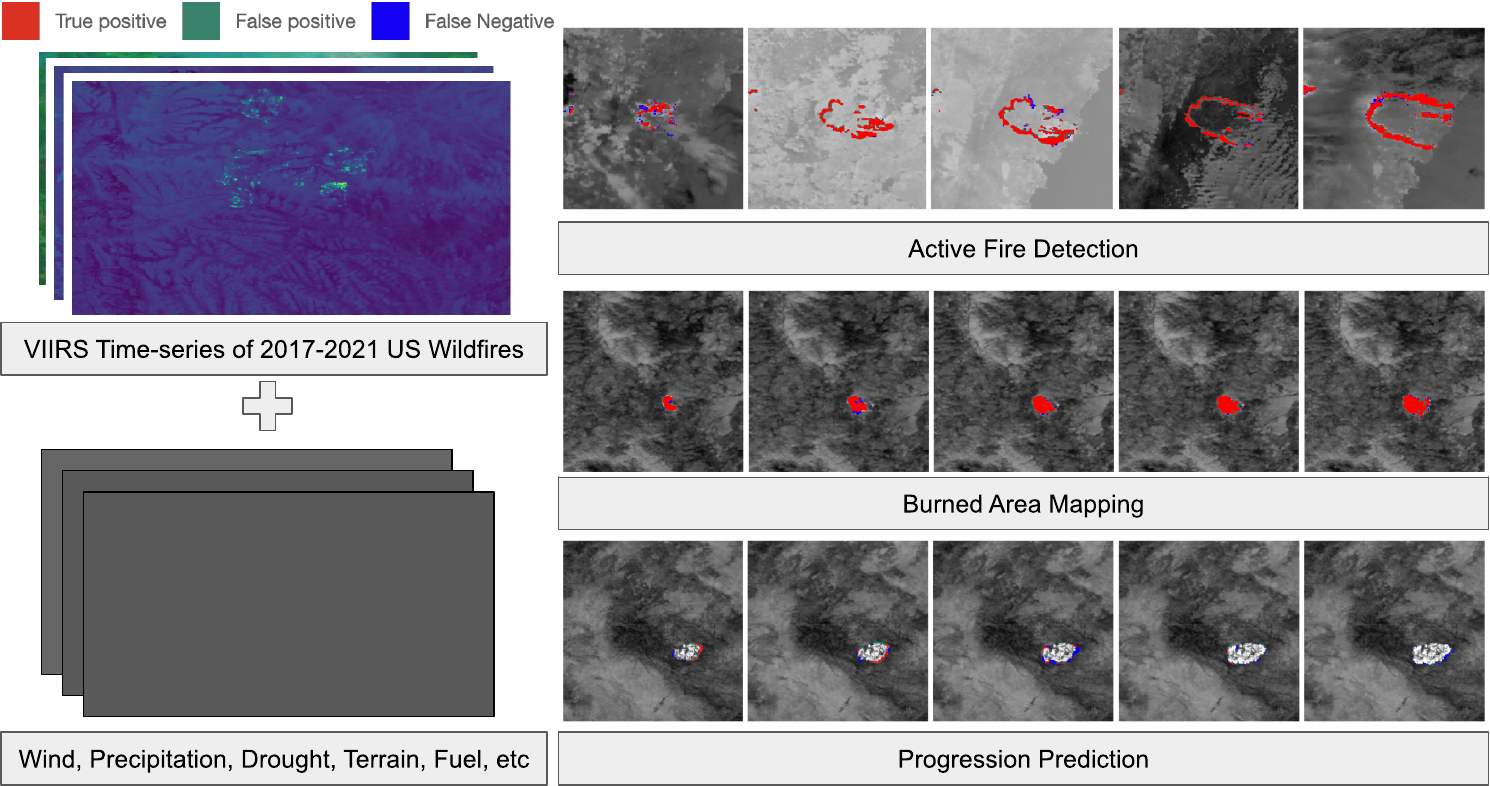}
    \caption{Major components of the TS-SatFire dataset and three distinct tasks.}
    \label{fig:enter-label}
\end{figure}
The release of this dataset aims to help researchers build robust deep learning models that can accurately detect active fires and burned areas, as well as forecast fire progression. By leveraging remote sensing datacubes, we establish 1D pixel-based temporal models, 2D image-based spatial models, and 3D spatial-temporal models as baselines for three distinct tasks and cross-compare the performance of the same architecture on different tasks. By comparing different sets of models, we assess the performance of each architecture across various tasks to identify the most suitable architecture for both detection and prediction tasks.\par
\subsection*{Active Fire detection}
Active fire detection is commonly accomplished through Active Fire (AF) products released by the National Aeronautics and Space Administration (NASA) and the European Space Agency (ESA). These AF products are typically derived from sensors such as the Moderate Resolution Imaging Spectroradiometer (MODIS), Visible Infrared Imaging Radiometer Suite (VIIRS), and Sea and Land Surface Temperature Radiometer (SLSTR) onboarding sun-synchronous satellites. \cite{Justice2002TheMF} introduces MODIS AF product with a contextual algorithm and updated in \cite{Giglio2003AnEC} by changing the threshold according to a larger context window. Currently, the MODIS Collection 6 active fire detection algorithm is still operating \cite{col6}. SLSTR onboarding Sentinel-3 also introduces its AF product in \cite{Xu2020FirstSO} which provides a temporal and spatial resolution like the MODIS AF product. While all AF products from these sensors offer a twice-daily temporal resolution, VIIRS holds an advantage in spatial resolution, providing 375 meters compared to the 1000 meters offered by MODIS and SLSTR \cite{Oliva2015AssessmentOV}. VIIRS AF products are generated from VIIRS images using a multi-spectral contextual method that involves multi-criteria thresholds \cite{Schroeder2014TheNV}. However, significant flaws with the current AF product include false positive detections resulting from cloud cover or buildings with high-temperature roofs, as indicated by \cite{Schroeder2014TheNV}. In recent years, deep learning models have shown success in remote sensing applications \cite{Zhu2017DeepLI}. For deep learning models employed in active fire detection, some utilize the rich temporal information of satellite image time-series, while others leverage spatial information to detect active fires within each image. In \cite{de2021active}, the authors published an active fire dataset based on mid-resolution Landsat-8 imagery and showed that ConvNets can effectively detect active fires. In \cite{Zhao2022GOESRTS}, Recurrent Neural Networks are shown to effectively process the time-series of coarse-resolution geostationary satellite images and detect early active fires. In \cite{Zhao2023TokenizedTI}, a Transformer-based model is proposed to process time-series of VIIRS images to detect active fires. It also indicates that time-series models have more advantages in detecting burned areas than using Conv-Net-based models. 
\subsection*{Burned Area Mapping}
Existing burned area products provided by NASA are often based on MODIS (MCD45 and MCD64) \cite{Roy2002BurnedAM, Roy1999MultitemporalAB, Giglio2009AnAB}. MCD45 relies on the reflectance difference between burned and unburned areas in the Short-wave Infrared (SWIR) and Near Infrared (NIR) bands. MCD64 further incorporates the Medium Infrared (MIR) band along with the SWIR-NIR difference. Additionally, there is a VIIRS-based burned area mapping product (VNP64A1), which provides a spatial resolution of 500 meters. However, these products only offer burned area information on a monthly basis. Daily mapping of wildfire progression remains a challenge. Another approach for burned area mapping involves clustering the hotspots detected by AF products each day \cite{Oliva2015AssessmentOV}. However, \cite{Scaduto2020SatelliteBasedFP} suggests that the accumulation of hotspots may lead to sparse burned area mapping which has significant omission errors. The Fire CCI product is introduced by the European Space Agency in \cite{LizundiaLoiola2020ASA, Chuvieco2018GenerationAA}. The product relies on the Near Infrared (NIR) band of MODIS to map burned areas on a monthly basis. However, the accuracy is also compromised by only using the NIR band. There are also multiple deep-learning models used for burned area mapping. Deep Learning for burned area mapping with mid-resolution sensors onboarding satellites like Sentinel-1/2 and Landsat-8/9 is broadly investigated \cite{Ban2020NearRW, Zhang2023TotalvariationRU, Zhao2022GlobalSB, Knopp2020ADL, Pereira2021ActiveFD}. However, due to their low temporal resolution, these methods are not suitable for monitoring burned areas at high frequency. With sensors possessing high temporal resolution, several deep learning models have been applied for continuous burned area mapping. For instance, \cite{Gmez2011PrototypingAA} employs a Multi-Layer Perceptron (MLP) for pixel-wise burned area mapping using four spectral bands of VIIRS images. Similarly, \cite{PINTO2020260} utilizes ConvLSTM to process VIIRS image time-series consisting of 750m spatial resolution moderate bands. This indicates that leveraging both spatial and temporal information can more effectively detect burned areas from VIIRS datacubes. MesogeosAM \cite{Kondylatos2023MesogeosAM} introduces a multi-task wildfire dataset focusing on the Mediterranean region. The dataset can also train deep learning models for burned area mapping tasks. However, the dataset is based on MODIS data which only provides 1000m spatial resolution.
\begin{table*}[!hbt]
\centering
\caption{Channels of the dataset and description of each channel}
\label{tab:bands}
\begin{tabular}{|c|c|c|c|c|c|} 
\hline
\textbf{Channel} & \textbf{Description} & \textbf{Channel} & \textbf{Description}     & \textbf{Channel} & \textbf{Description}         \\ 
\hline
1                & Band I1              & 10               & EVI                      & 19               & Aspect                       \\
2                & Band I2              & 11               & Total Precipitation      & 20               & Elevation                    \\
3                & Band I3              & 12               & Wind Speed               & 21               & PDSI                         \\
4                & Band I4              & 13               & Wind Direction           & 22               & Land Cover                   \\
5                & Band I5              & 14               & Min Tempreture           & 23               & Total Precipitation Surface  \\
6                & Band M11             & 15               & Max Tempreture           & 24               & Forecast Wind Speed          \\
7                & Band I4 Night        & 16               & Energy Release Component & 25               & Forecast Wind Direction      \\
8                & Band I5 Night        & 17               & Specific Humidity        & 26               & Forecast Tempreture          \\
9                & NDVI                 & 18               & Slope                    & 27               & Forecast Specific Humidity   \\
\hline
\end{tabular}
\end{table*}
\subsection*{Wildfire Progression Prediction}
Commonly used methods for predicting wildfire progression, like FARSITE \cite{finney_farsite_1998} and Prometheus \cite{tymstra_development_2010}, are semi-empirical methods, combining physical knowledge with some aspects determined from experimental observations. In recent years, purely empirical approaches based on deep learning models have been investigated as an alternative. FireCast \cite{Radke2019FireCastLD} utilizes a 6-layer Convolutional Neural Network and Landsat-8 images together with elevation, weather and wind information to forecast the progression on the next day. The model shows better accuracy compared to FARSITE. In \cite{singla_wildfiredb_2021}, a tabular dataset is proposed to predict if the fire propagates to neighbourhood locations based on weather, elevation, and vegetation information at that location. NextDayWildfireSpread \cite{huot_next_2022} provides an image-based dataset consisting of information related to weather, elevation, vegetation and population density. It proposes to use a convolutional autoencoder to process the input data to forecast the active fire on the next day. Following NextDayWildfireSpread, WildfireSpreadTS \cite{Gerard2023WildfireSpreadTSAD} leverages multi-temporal information for wildfire progression prediction, compared to the mono-temporal approach in \cite{huot_next_2022}. Besides most of the auxiliary data modalities used by NextDayWildfireSpread, WildfireSpreadTS adds land cover information and weather forecasts. Moreover, the active fire masks are improved from MODIS to VIIRS, which provides a better spatial resolution. WildfireSpreadTS uses U-Net, ConvLSTM and UTAE as baseline models, though baseline results highlight the difficulty of the progression prediction task. Notably, research integrating wildfire monitoring tasks, such as active fire and burned area detection, with wildfire progression prediction remains limited. With the advent of foundation models, multi-task datasets like this can serve as both training and benchmark datasets, playing a pivotal role in advancing Earth observation foundation models.

\section*{Methods}

\label{dataset_section}
The details of the spectral bands and auxiliary data are provided in this section. VIIRS images are downloaded from NASA's Level-1 and Atmosphere Archive \& Distribution System (LAADS) and processed locally. All the auxiliary data are downloaded and processed by Google Earth Engine \cite{gee}. 
\section*{Spatial distribution of fires}
\label{firesmaps}

\begin{figure}[!hbt]
  \centering
  \begin{subfigure}{.7\linewidth}
    \includegraphics[width=\linewidth]{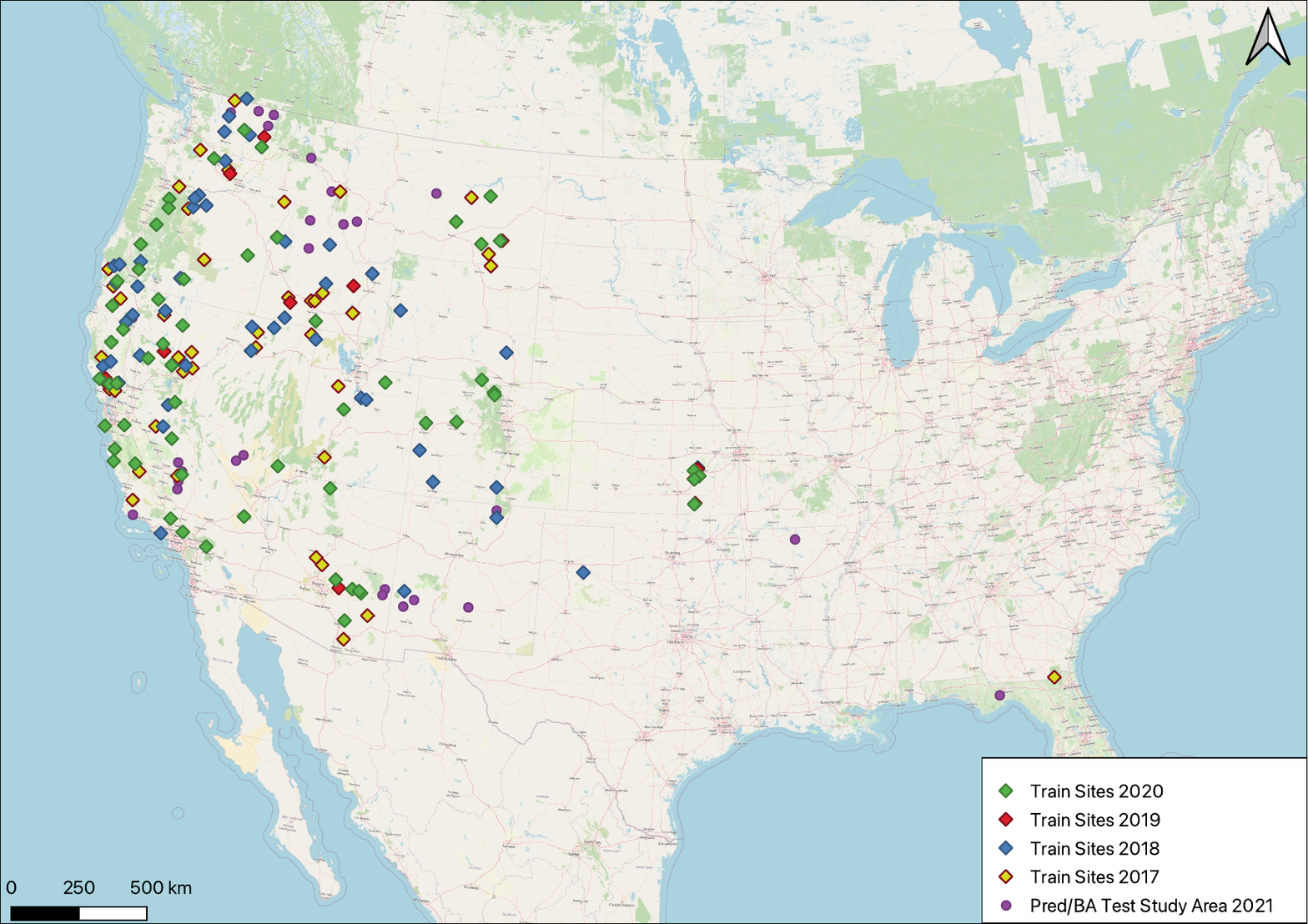}
    \caption{Training Dataset for all tasks and Test Study Areas for Burned Area Mapping and Fire Progression Prediction task}
    \label{fig:image1}
  \end{subfigure}\\
    
  \begin{subfigure}{.7\linewidth}
    \includegraphics[width=\linewidth]{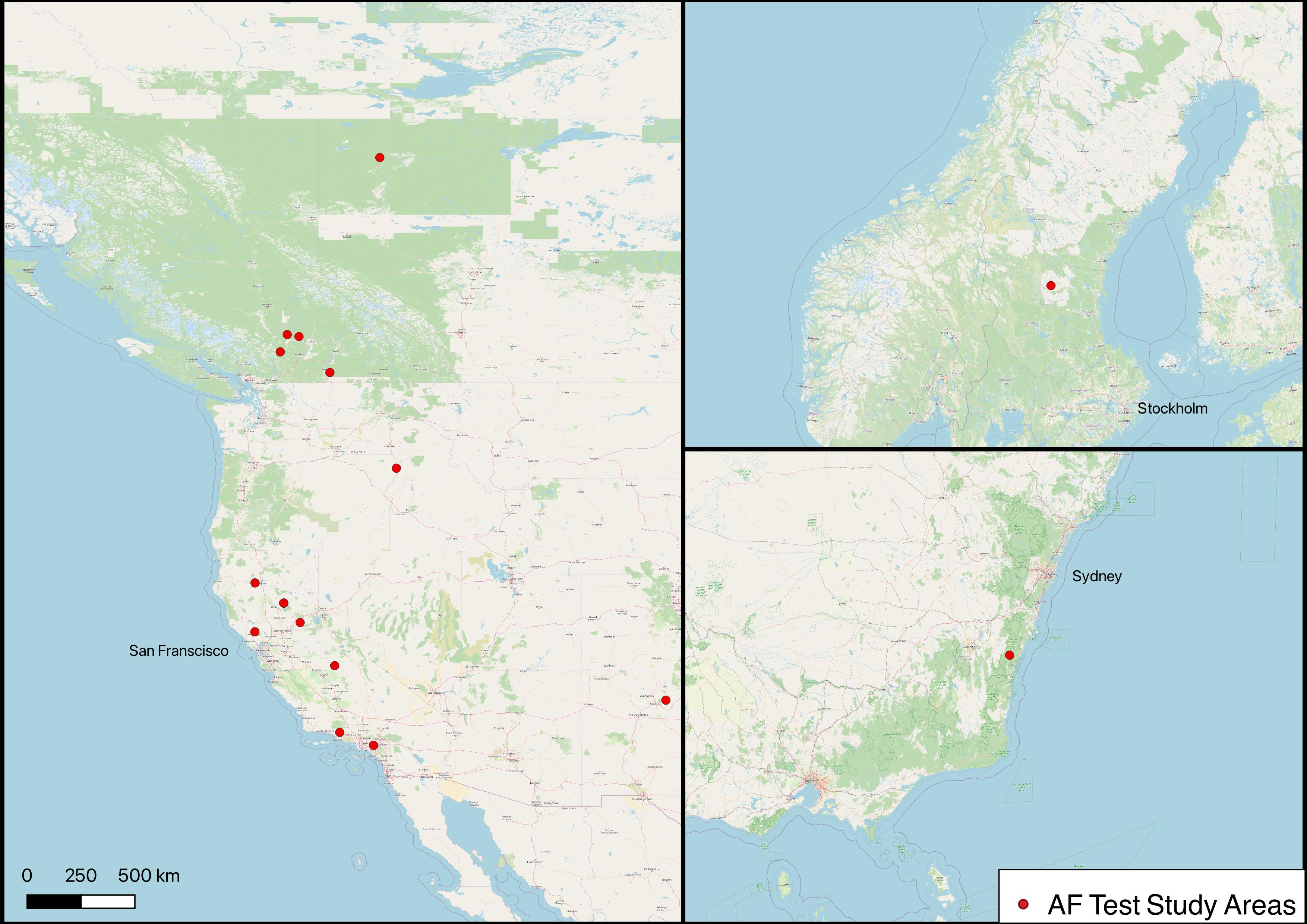}
    \caption{Test Study Areas for Active Fire detection task}
    \label{fig:image3}
  \end{subfigure}
  \caption{Overview of locations and land covers of fires in an example split into train/test/validation set. The legend in the test set does not cover any fires.}
  \label{fig:map}
\end{figure}

As shown in Figure \ref{fig:map}, the training dataset consists of wildfire events between 2017 and 2020 covering the US main continent. In total, there are 34 fire events used in 2017, 34 events in 2018, 8 events in 2019 and 49 events in 2020. As for the test set of burned area mapping and fire progression prediction tasks, 24 fire events from 2021 are used. There are 13 wildfire events picked from 2017-2020 used as the validation set. For active fire detection tasks, the test set consists of 17 wildfire events between 2018 and 2022 across multiple continents.

\subsection*{Overview of the input data sources}
\begin{table*}[!hbt]
\centering
\caption{Overview of Data Sources Used as Input for Burned Area Detection, Active Fire Detection and Progression Prediction Tasks.}
\label{tab:features-overview}
\begin{tabular}{lll} 
\toprule
Original dataset & Feature                                                                                                                   & Resolution      \\ 
\midrule
VNP02IMG         & VIIRS/NPP Imagery Resolution 6-Min L1B Swath 375m                                                                         & 375m / 12h      \\
VNP03IMG         & \begin{tabular}[c]{@{}l@{}}VIIRS/NPP Imagery Resolution Terrain-Corrected \\Geolocation 6-Min L1 Swath 375m\end{tabular}  & 375m / 12h      \\
VNP02MOD         & VIIRS/NPP Moderate Resolution 6-Min L1B Swath 750m                                                                        & 750m / 12h      \\
VNP03MOD         & \begin{tabular}[c]{@{}l@{}}VIIRS/NPP Moderate Resolution Terrain-Corrected \\Geolocation 6-Min L1 Swath 750m\end{tabular} & 750m / 12h      \\
GlobFire         & GlobFire wildfire events and fire perimeters based on MODIS                                                               & 500m / Monthly  \\
MCD64A1          & MODIS Monthly Burned Area Product                                                                                         & 500m / Monthly  \\
SRTMGL1 v003~    & Elevation (derived from this: aspect, slope)                                                                              & 30m             \\
MCD12Q1.061~     & Land cover class                                                                                                          & 500m            \\
GRIDMET~         & \begin{tabular}[c]{@{}l@{}}Wind, rain, temperature, humidity, drought index, \\energy release component\end{tabular}      & 4638.3m / 24h   \\
NOAA\_GFS0P25~   & Forecast for wind, rain, temperature, humidity                                                                            & 27830m / 1h     \\
\bottomrule
\end{tabular}
\end{table*}

\subsection*{VIIRS Imagery}
\label{viirs_imagery}
The VIIRS sensor is operational on multiple satellites, including Suomi-NPP, NOAA-20, and NOAA-21. This sensor provides data across 22 different spectral bands, categorized into Imagery Bands and Moderate Bands. The Imagery Bands cover wavelengths from Red to Long-wave Infrared, offering a spatial resolution of 375 meters at nadir. The Moderate Bands cover a broader range of wavelengths but have a spatial resolution of 750 meters. In TS-SatFire, six spectral bands are utilized, including Imagery Bands I1-I5 and Moderate Band M11, as shown in Table \ref{tab:bands}. Band I4, the Medium Infrared Band with a wavelength of 3.7 $\mu$m, is particularly effective for detecting active fires due to its alignment with the black-body radiation curve peak of forest fires. Similarly, Long-wave Infrared Band I5 is used to detect higher-temperature fires. Both Bands I4 and I5 provide two captures daily: one during the daytime and another at night. Near Infrared Band I2 and Short-wave Infrared Bands I3 and M11 are employed to visualize burned areas by leveraging the reflectance differences between burned and unburned regions. The Red Band I1 is essential for distinguishing clouds and smoke due to its shorter wavelength. This combination of spectral bands ensures effective monitoring and analysis of wildfire events. 
\subsection*{Auxiliary Data}
For the progression prediction task, the auxiliary data follow the setup of WildfireSpreadTS. As shown in Table \ref{tab:bands}, there are three sub-classes of these data, weather/weather forecast, topography, and landcover.
\subsubsection*{Weather Data and Forecast Data}

The Gridded Surface Meteorological Dataset (GRIDMET) \cite{abatzoglou_development_2013} and the Global Forecast System (GFS) \cite{clough_atmospheric_2005} supply the weather data and its forecast, respectively, and are exclusively available for the US region. GRIDMET offers a spatial resolution of 4638 meters, providing a coarse observation of the weather. Since wind plays a significant role in wildfire propagation, factors such as wind speed and direction are prioritized. Additionally, precipitation, specific humidity, and the Palmer Drought Severity Index (PDSI) influence vegetation and soil moisture, consequently impacting wildfire ignition negatively \cite{Littell2016ARO}. Extreme temperatures have also been closely correlated with wildfire events \cite{Brown2023ClimateWI}, hence temperature data is included in this dataset. Furthermore, the Energy Release Component, an index proposed by the National Fire Danger Rating System (NFDRS) that describes the potential intensity of a wildfire, is incorporated. On the other hand, GFS offers hourly weather forecasts at a spatial resolution of 27.83 kilometers. It forecasts wind, temperature, and humidity information on an hourly basis, akin to GRIDMET. To synchronize the temporal resolution, the dataset utilizes the average value over 24 hours of forecast data, based on forecasts made at the end of the 'current' day, without using future information. 

\subsubsection*{Topography and Landcover Data}
Topography can affect the wildfire progression by affecting the wind flow\cite{Povak2018EvidenceFS}. We use the NASA SRTM Digital Elevation dataset \cite{nasa_jpl_nasa_2013} to derive the slope and aspect of the surface. The SRTM dataset provides 90m spatial resolution, and SRTM version 3 is used in the dataset. Land cover information is essential to understand the fuel types of the burning biomass. Fuel types affect the progression speed and severity of the wildfire. The land cover used in this project is based on the 500m MODIS Land Cover Type Yearly Global product (MCD12Q1.061)\cite{sulla-menashe_hierarchical_2019,friedl_modisterraaqua_2022}. 

\subsection*{Labels}
\label{used_labels}
\subsubsection*{Active Fire Label}
For the active fire detection task, the training labels are sourced from the NASA VIIRS AF product. Due to potential errors within the AF product, we manually inspect the AF labels visually as a quality control procedure, ensuring that the AF labels correspond to the bright spots observed on Band I3-I5 and M11. Areas that do not pass this inspection are removed from the training set. For the test labels, we manually set the threshold to Band I4/I5 to ensure alignment with the bright spots observed in the images. Examples are included in the supplementary material for reference. In total, 17 wildfire events across the globe are utilized as the test sites, with geolocations and start/end dates of each test wildfire event provided in the supplementary material.

\subsubsection*{Burned Area Label}
Creating the burned area labels is challenging because they require daily captures. While this is also true for active fire labels, the spectral bands which are sensitive to the burned area (Short-wave Infrared and Near-Infrared) are more prone to be affected by clouds and smoke than those used to detect active fire (Mid-Infrared). We use two sources of labels: accumulation of VIIRS AF product, and daily burned area perimeters from the National Interagency Fire Center (NIFC). NIFC polygons are generated from mixed methods, including fieldwork and extraction from aerial images. However, as mentioned in Section 2, the accumulation of AF points results in omission errors. Also, NIFC polygons have very low accuracy at the beginning of the wildfire and there are high commission errors in some study areas. Therefore, the union of the accumulated VIIRS AF detections and the NIFC perimeters is used as the training label. For fires used in the test set, we either a) determine the labels in the same way as for the training set, or b) we only use the accumulated VIIRS AF detections, to avoid the high commission errors of NIFC perimeters. We determine whether to choose A or B by visually inspecting which approach better covers the burned area highlighted in Short-wave Infrared, Near-Infrared or Mid-Infrared bands. This is an additional quality control that we only use on the test set.

\subsubsection*{Progression Prediction Label}
For the progression prediction task, the labels are defined as the difference between the burned area masks of the last observed day and the next day. 
A naive approach would be to use the raw burned area mask of the next day as the label. However, this would obfuscate the model's real prediction capabilities, since simply predicting the last day's burned area would already yield a very high performance. As the input data already includes VIIRS bands that correlate strongly with burned area, the model might actually just learn to perform burned area detection on the last observed day, instead of predicting the next day's burned area. To avoid this, we let the model only predict the newly burned area on each day, by taking the difference between subsequent burned area labels. 
\subsection*{Preprocessing}
For active fire detection, the arrays from the GeoTIFF file are directly used as input after the normalization. For the burned area mapping task, bands I4 and I5 in the day and night captures are aggregated by taking the pixel-wise maximum over all images in the current and previous timestamps. This is because Band I4 and I5 are sensitive to the ground temperature. By aggregating all previous images with the maximum value, total burned areas will be highlighted in these two spectral bands. The same preprocessing of Band I4-I5 is also applied to spectral bands of the fire prediction tasks. As for the auxiliary data, the pixel-wise median of GRIDNET weather data and the pixel-wise mean of weather forecast data are used. For the drought index (PDSI) and Landcover, the pixel-wise median data are used.\par
To go from the full time-series of images for each fire to a fixed-size input that we can present to the various models, we first sample the image time-series with a length of $T$ from the full image time series of each fire. For active fire detection and burned area mapping, the sampling interval between each window is $1$ for the training set and $T$ for the test set. For the prediction task, the sampling interval for the testset is set to $1$. Since different models expect different input shapes, further processing is applied. Temporal models like GRU, LSTM and T4Fire take pixel time-series as the input. Therefore, each image time-series with shape $(W, H, T, C)$ is divided into $W * H$ pixel time-series with the shape $(T, C)$. For the spatial models, each image within the image time-series is used as an individual sample. For the spatial-temporal model, the image time-series with length $T$ is directly used as the input.

Data sources used in this work are presented in Table \ref{tab:features-overview}. Satellite imagery is generated from four Level-1B VIIRS satellite products. VNP02IMG and VNP02MOD provide the raster of 375m imagery bands and 750m moderate bands and VNP03IMG and VNP03MOD provide the geolocation of the raster. For other auxiliary data, they are resampled to 375m with bilinear interpolation.

\subsection*{Missing values}
\label{misval}
\begin{figure}
    \centering
    \includegraphics[width=1\linewidth]{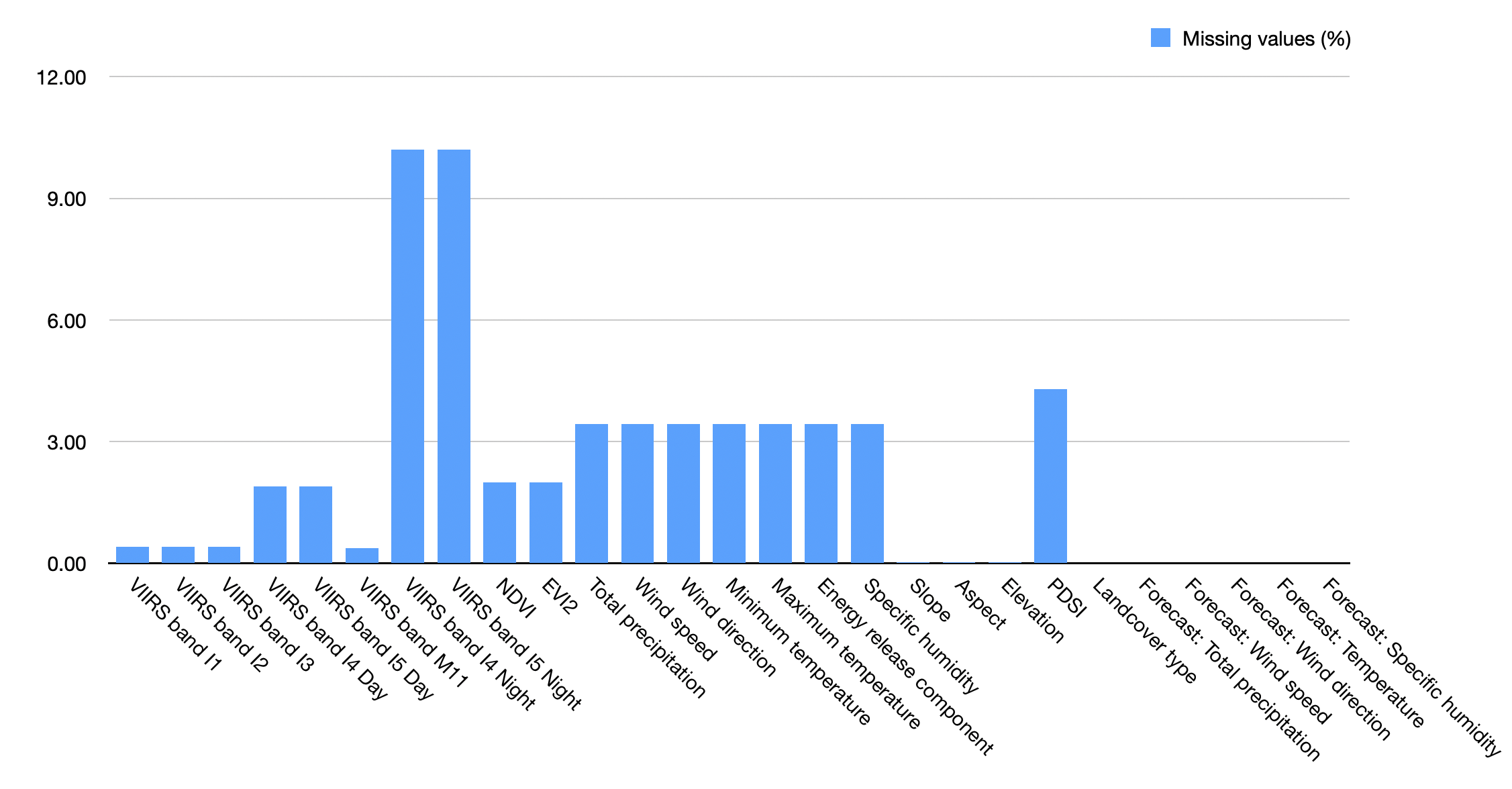}
    \caption{Percentage of missing values: Rates of missing values of each spectral band used in active fire detection, burned area mapping and fire progression prediction tasks. All the missing values are replaced with zeros during training and testing.}
    \label{fig:missing-values}
\end{figure}

The percentage of missing values is shown in Table \ref{fig:missing-values} for all the features used in active fire detection, burned area mapping and fire progression prediction tasks. The rates of the six spectral bands of day images are below 2$\%$ of measurements. Rates of spectral bands of night images have around 10$\%$ missing values, which indicates the night images are not always available. For other auxiliary data, there rates of missing value are below 5$\%$. In the preprocessing, all the missing values are replaced with zeros.
\section*{Data Records}
The TS-SatFire dataset is available on Kaggle (\url{https://www.kaggle.com/datasets/z789456sx/ts-satfire})\cite{yu_zhao_2024}. It includes 179 distinct wildfire events, each organized into a separate folder named by its fire ID. Fire IDs are derived from the GlobFire dataset (training set 2017–2020), the MODIS monthly burned area product (BA/Pred test set 2021), and corresponding names (AF test set). Within each folder, auxiliary data are stored in the FirePred folder, while VIIRS images are separated into VIIRS\_Day and VIIRS\_Night folders. All data are provided in GeoTIFF format and are co-registered to the same region of interest. The dataset captures the lifecycle of each wildfire, covering events in the contiguous U.S. from January 2017 to October 2021. It comprises 3552 surface reflectance images along with auxiliary data such as weather, topography, land cover, and fuel information, amounting to a total size of 71 GB.
\section*{Technical Validation}
\label{sec:method}
\subsection*{Candidate Temporal Models for Benchmark}
Purely temporal models are only used in active fire detection tasks. These models are designed to classify pixel time-series. The results generate an active fire map according to the input pixels' original positions.\par
\textbf{GRU/LSTM} For sequential models like GRU and LSTM, we follow the model setup used in \cite{Zhao2023TokenizedTI}. Both GRU and LSTM models have a hidden size of 64 and consist of 3 layers each. A dense layer is then used to classify each pixel as fire or non-fire.\par
\textbf{T4Fire} T4Fire is a Transformer-based model proposed in \cite{Zhao2023TokenizedTI} for classifying pixel time-series. We adhere to the same hyperparameters as proposed in \cite{Zhao2023TokenizedTI}.\par

\subsection*{Candidate Spatial and Spatial-Temporal Models for benchmark}
All spatial models are applied to both active fire detection and burned area detection tasks. These models take 2-dimensional images as input and produce the corresponding active fire and burned area maps as output.\par
\textbf{U-Net} We use a U-Net~\cite{unet} as the weak baseline for both detection tasks, using only 2D input. For the U-Net-3D, the 2D convolutions are exchanged with 3D convolutions. The model thus treats the temporal dimension similarly to the two spatial dimensions, except that we do not use downsampling in the temporal dimension in any of our models.\par
\textbf{Attention U-Net} Attention U-Net is an improved version of the U-Net. It uses an attention mechanism, which enables learning with a focus on specific regions of the image. Similarly, Attention U-Net-3D also changes 2D convolutions to 3D to process 3D tensors.\par
\textbf{UNETR} UNETR \cite{hatamizadeh2021unetr} has a similar shape as U-Net. The major difference is the encoder, which is changed from a ConvNet to a Vision Transformer. The model was originally proposed to process 3D data. To process 2D data with it, we use a variant UNETR-2D, which is only applied to the spatial dimensions as a U-Net. For time-series input, the model treats the temporal dimension as the third spatial dimension, in addition to height and width.\par
\textbf{SwinUNETR} SwinUNETR \cite{Tang2021SelfSupervisedPO} was designed for 3D semantic segmentation tasks, such as those involving 3D medical images. Similar to UNETR and U-Net, it has an encoder-decoder architecture with skip connections. The major difference is the backbone, which is changed from a Vision Transformer to a Swin-Transformer \cite{Liu2021SwinTH}. For 2D input, SwinUNETR processes only the image instead of the time-series. For time-series input, the patches are divided only within the spatial dimensions, enabling attention across different timestamps. Additionally, the attention windows in the Swin Transformer blocks span the full length of the time-series to ensure full temporal attention.\par

\begin{table*}[!hbt]
\centering
\caption{Quantitative results of AF and BA tasks of three sets of baseline models: spatial models, temporal models and spatial-temporal models.}
\label{quant}
\begin{tabular}{cccccc} 
\toprule
\textbf{Model Name} & \textbf{\#Params} & \textbf{Task}  & \textbf{F1}                    & \textbf{IoU}                   & \textbf{Input Days}  \\ 
\hline
\multicolumn{6}{l}{\textbf{Spatial models }}                                                                                                      \\ 
\hline
U-Net               & 10.6M             & AF / BA        & 0.731 / 0.789                  & 0.605 / 0.676                  & 1                    \\
Attention-U-Net     & 31.7M             & AF / BA        & 0.763 / 0.793                  & 0.648 / 0.687                  & 1                    \\
UNETR-2D            & 23.52M            & AF / BA        & 0.733 / 0.837                  & 0.621 / 0.750                  & 1                    \\
SwinUNETR-2D        & 25.2M             & AF / BA        & 0.774 / 0.829                  & 0.66 / 0.733                   & 1                    \\ 
\hline
\multicolumn{6}{l}{\textbf{Temporal models }}                                                                                                     \\ 
\hline
GRU-3               & 64.3K             & AF             & 0.713 / -                      & 0.601 / -                      & 6                    \\
LSTM-3              & 84.9K             & AF             & 0.765 / -                      & 0.654 / -                      & 6                    \\
T4-Fire             & 32.5K             & AF             & 0.802 / -                      & 0.700 / -                      & 6                    \\ 
\hline
\multicolumn{6}{l}{\textbf{Spatial-Temporal models }}                                                                                             \\ 
\hline
U-Net-3D            & 31.7M             & AF / BA~/ Pred & 0.748 / 0.834 / \textbf{0.375}          & 0.628 / 0.746 / \textbf{0.338}          & 6                    \\
Attention-U-Net-3D  & 94.5M             & AF / BA~/ Pred & 0.770 / 0.712 / 0.354          & 0.654 / 0.607 / 0.312          & 6                    \\
UNETR-3D            & 34.8M             & AF / BA~/ Pred & \textbf{0.811} / 0.823 / 0.371 & \textbf{0.706} / 0.736 / 0.336 & 6                    \\
SwinUNETR-3D        & 33.2M             & AF / BA~/ Pred & 0.797 / \textbf{0.855 }/ 0.374 & 0.688 / \textbf{0.768 }/ 0.331 & 6                    \\
\bottomrule
\end{tabular}
\end{table*}
\label{experiment section}
For AF and BA tasks, three classes of models are tested in the main paper: temporal, spatial, and spatial-temporal. For the temporal models, since the model is from the same task with the same data source, the same hyperparameter as the original paper \cite{Zhao2023TokenizedTI} is used for three candidate models, GRU, LSTM and T4Fire. GRU and LSTM have 64 as their hidden size, and there are three layers of GRU and LSTM for each candidate model. Finally, there is a dense layer to classify features into 2 classes. For the Transformer blocks of the T4Fire model, the number of heads is set to 3, the embedding size is set to 24, and the size of the dense layer in the Transformer block has a hidden size of 112. There are in total 4 layers of Transformer blocks in total. The training is completed with a batch size of 1024, a learning rate of 0.0003 and focal cross-entropy loss. The large batch size is because temporal models are processing time-series of pixels. U-Net and Attention-U-Net follow the default implementation, which has [64, 128, 256, 512, 1024] as the intermediate feature size with a stride of 2. For UNETR and SwinUNETR, they are originally proposed to process three-dimensional data. UNETR-2D and SwinUNETR-2D reduce the input dimension and divide the input 2D images into 2D patches instead of 3D. UNETR-3D and SwinUNETR-3D are both adopted to process the image time-series by dividing the image time-series only along spatial dimensions while applying attention across both spatial and temporal dimensions. This is because the size of the temporal dimension is significantly lower than the spatial dimension. The patch size is set to (1, $P$, $P$), where $P$ is the width and height of an image patch. For UNETR-2D and UNETR-3D, $P$ is set to its default value of 16, for SwinUNETR, the patch size is set to 2. The size of the attention window of SwinUNETR-3D is set to ($T$, 4, 4) which T denotes the length of the time-series. For the hyperparameter search of UNETR and SwinUNETR, the feature size is tested with [24, 36], and 36 is picked in the end with a better F1 Score and IoU Score. Features larger than 36 are not tested because of the hardware limitations. For UNETR, the hidden size is set to 384 and the MLP size is set to 1536 to ensure it has a similar number of parameters as U-Net. In the end, SwinUNETR, UNETR, and U-Net all have around 30M parameters to keep the fairness of the comparison. The training is completed with a batch size of 4, a learning rate of 0.0003 and DiceLoss.\par
For the prediction task, the models are trained with hardware with more VRAMS, which enables larger batch sizes than other tasks. Also, the loss function is switched to DiceCELoss since the models were not able to escape from this local minimum when trained with DiceLoss. We assume that this is caused by the class imbalance in the data. We set its weight parameter to [1.0, 446.7836] to up-weight the burned area class according to the class distribution in the training set. We performed a short grid search over using a 1:1 mixture of Dice and CE loss vs. using only CE loss, combined with learning rates in \{0.01, 0.001\}. Using this grid search, we were able to find hyperparameters for which we could avoid the local minimum. For U-Net-3D, UNETR-3D and SwinUNETR-3D, we used a 1:1 mixture of Dice and CE loss with a learning rate of 0.001, with batch sizes of 64 for the U-Net-3D, 4 for the UNETR-3D, and 8 for the SwinUNETR-3D. The Attention-U-Net-3D did not end up in the local minimum in the initial hyperparameter search, so for it, we kept the original setup of using Dice loss, the best learning rate of 0.0012 and a batch size of 8. \par

\subsection*{Baseline Results}
\label{baselinereuslts}

The quantitative baseline results are reported in Table \ref{quant} and the qualitative results are provided in the supplementary material. For the AF task, all models are tested over 17 wildfire events across the globe. The geolocation of these study areas can be found in the supplement material. Among all models for AF detection, T4Fire can achieve better prediction results compared to other spatial models and temporal models with a significantly lower number of parameters. Noticeably, all temporal models have a lower number of parameters. This is because the solution is pixel-based. Consequently, the inference speed is the major bottleneck of the methods. The spatial-temporal model UNETR-3D achieves a slightly better performance than T4Fire. Compared with UNETR-2D, which has lower performance in the AF task, the result of UNETR-3D highlights the importance of using temporal information to detect active fire.\par
For the BA task, the temporal models have not been tested. The test dataset consists of 24 wildfire events generated from 2021 wildfire events in the US. Comparing spatial-temporal models with spatial models on burned area tasks, spatial-temporal models like SwinUNETR-3D show privilege in F1 Score and IoU Score over all other spatial models including the 2D version of SwinUNETR. It demonstrates the importance of using temporal information when segmenting low-resolution satellite images which also agrees with the finding of the AF task.\par
Comparing the results of spatial models between the AF and BA tasks, all results from the BA task are higher than the AF task. One major reason for this is the test datasets are different. However, from the comparison between spatial and temporal models, it can be observed that spatial models are not suitable for detecting active fires. The main reason is that AF labels are generally more sparse than BA labels, which makes the spatial information less useful in detecting AF than BA.\par

In the spatial-temporal models section in Table \ref{quant}, the quantitative results for burned area mapping and progression prediction tasks are compared. The prediction F1 Scores and IoU Scores are notably lower than those of the detection tasks, reflecting the greater complexity of the prediction task, which cannot be effectively addressed by standard segmentation models "out of the box." While U-Net-3D achieves the best performance among the baseline models, the quantitative results across all four baselines remain similar. This highlights that image segmentation models like U-Net and SwinUNETR, designed to extract spatial features from input data, are not well-suited for prediction tasks that require inferring future states from past information.
\begin{table}[!hbt]
\centering
\caption{Variations of the F1 Score and IoU score according to different random seeds. Three different random seeds (42, 43, 44) are used to train the model with the best performance of each task. The mean and standard deviation of models with different random seeds are reported in this table. The result of the AF task is from UNETR-3D, results of the BA and Pred tasks are from SwinUNETR-3D.}
\label{tab:variations}
\begin{tabular}{cccc} 
\toprule
\textbf{Metrics} & \textbf{AF} & \textbf{BA} & \textbf{Pred}  \\
\hline
F1            & 0.814±0.011                    & 0.853±0.007                        & 0.371±0.003                           \\
IoU           & 0.709±0.013                    & 0.767±0.009                        & 0.332±0.003                           \\
\bottomrule
\end{tabular}
\end{table}
As shown in Table \ref{tab:variations}, the models with the best F1 Score and IoU Score are trained with three different random seeds. The standard deviations of these models are $\leq$0.013 for the three tasks, indicating low variation due to the random initialization of the models.
\subsection*{Ablation Study}
\subsubsection*{Time-series Length}
\begin{table}[!hbt]
\centering
\caption{Ablation study on the length of the time-series, length equals 2, 4, 6 are tested for BA, AF and Prediction tasks. The tested model is SwinUNETR-3D.}
\label{tempab}
\begin{tabular}{ccccccc} 
\toprule
\textbf{Task}                        & \multicolumn{2}{c}{\textbf{TS=6}} & \multicolumn{2}{c}{\textbf{TS=4}} & \multicolumn{2}{c}{\textbf{TS=2}}  \\ 
\hline
     & F1    & IoU              & F1    & IoU              & F1    & IoU               \\
     \hline
BA          & 0.855 & 0.768             & 0.856 & 0.771            &     \textbf{0.869} & \textbf{0.789}        \\
AF                  &  0.783     &   0.667               &   0.803    & 0.703                 &    \textbf{0.823}   & \textbf{0.727}                  \\
Pred                  &     \textbf{0.374}               &  \textbf{0.331}        & 0.354        &  0.317                 &   0.366    &   0.321                \\ 
\bottomrule
\end{tabular}
\end{table}
For temporal models and spatial-temporal models, the length of the input time-series determines how much temporal information the model consumes. In Table \ref{tempab}, the lengths of the times-series are set to $2,4,6$ and the metrics under these configurations are provided. The F1 and IoU Scores of SwinUNETR-3D for the BA and AF tasks decrease as the time-series length increases, indicating that this naive adaptation does not effectively utilize the additional temporal information. However, compared to SwinUNETR-2D, SwinUNETR-3D shows improved F1 and IoU Scores when using a 2-day image time-series, suggesting some advantage in leveraging limited temporal context. For the prediction task, both F1 and IoU Scores increase as the time-series length grows, likely because forecasting benefits from longer input sequences that provide more comprehensive temporal context.\par
\begin{figure}
    \centering
    \includegraphics[width=\linewidth]{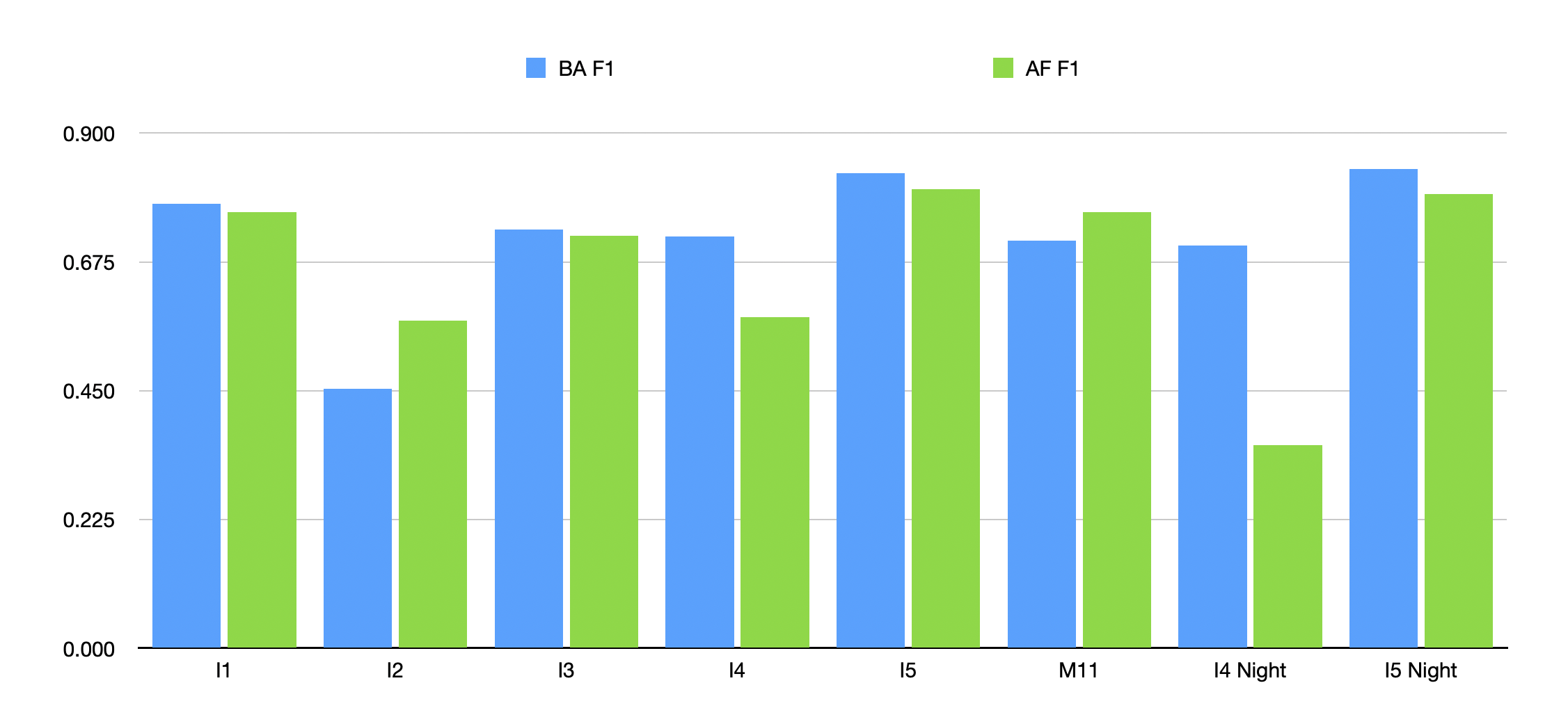}
    \caption{Feature Importance of the input bands for BA task (SwinUNETR-3D) and AF task (T4Fire).}
    \label{featureimport}
\end{figure}

\subsubsection*{Spectral Feature Importance}
In Figure \ref{featureimport}, we measure the importance of each feature by setting that feature to zero and assessing the resulting performance in AF and BA tasks. For the prediction task, the result is presented in the supplementary material. Band I2 contribute the most to the performance of the BA and AF tasks. The Near Infrared Band I2 contrasts the burned area and the active fire with the detection of I3, I4, and M11. Band I4 detects temperature anomalies which is crucial for detecting active fire. The burned area can be detected by aggregating the Band I4 images from the start of the wildfire based on their maximum value. Band M11 and Band I3 have similar contributions to the performance of AF and BA tasks. Short-wave Infrared bands M11 and I3 are useful in differentiating burned and unburned areas. For the thermal band I5, both day and night captures do not significantly contribute to the results.
\subsubsection*{Feature importance for fire prediction}

\begin{figure}
    \centering
    \includegraphics[width=\linewidth]{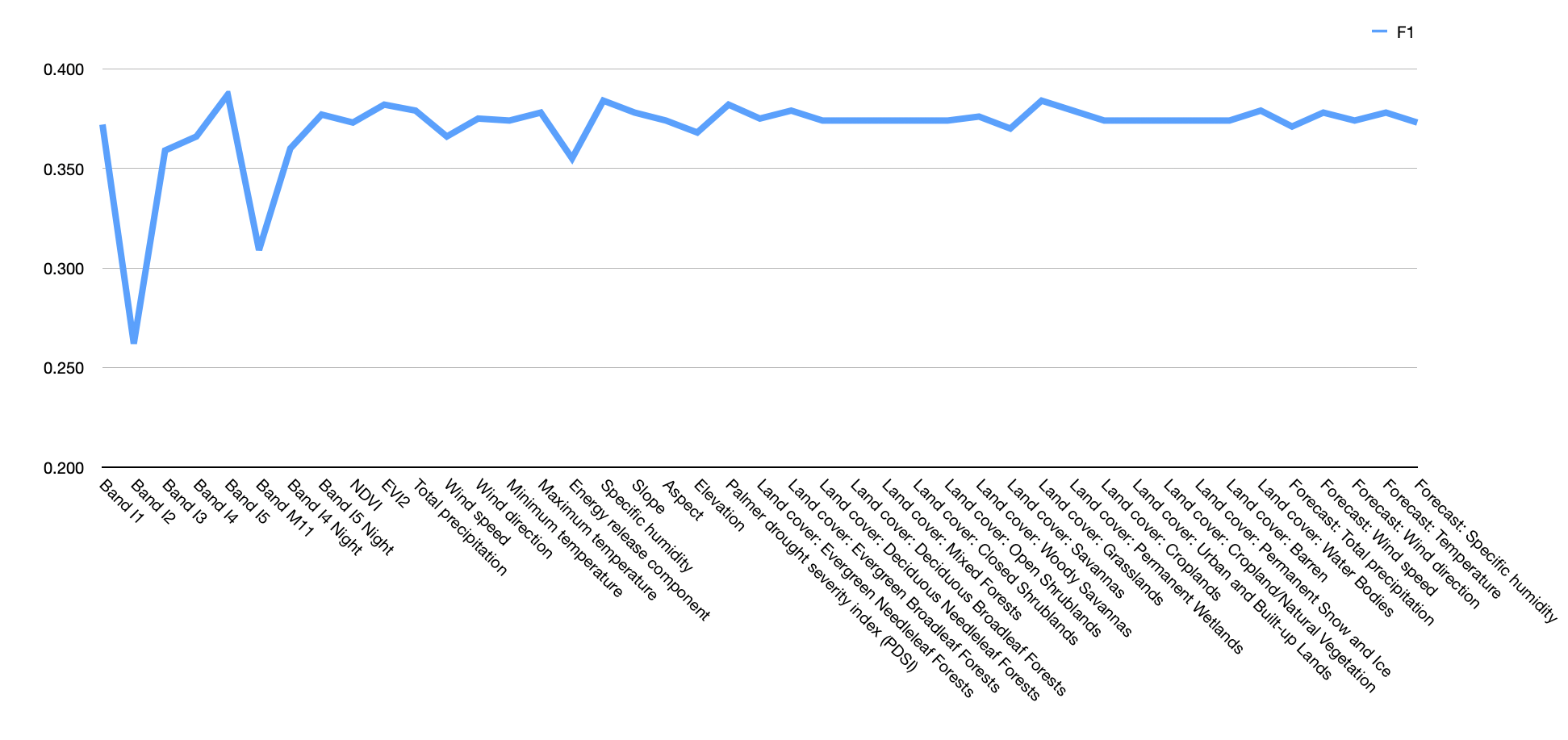}
    \caption{Feature importance for fire prediction task: We investigate a simple measure of feature importance, by setting one feature to zero in all inputs and measuring the resulting test F1 score. Removing important features should reduce the performance while removing unimportant features should have little influence.}
    \label{fig:feature-importance}
\end{figure}

As a very basic indicator of feature importance, we remove the information in one individual feature at a time by setting it to zero. For the standardized features, this represents setting it to their mean value. For the features that use degrees in [0,360], i.e. wind direction, forecast wind direction and aspect, this represents 0 degrees, and for the one-hot encoded land cover class, it represents an absence of all land covers. The results are shown in Figure \ref{fig:feature-importance}. They show that bands I2 and M11 are the most important features, leading to a reduction in F1 score of 11\% and 6.5\% when zero-ed out, respectively. In the main paper, band I2 also emerged as the most important band for the burned area detection, while band M11 was the third most important, though with much less influence than I2. Unlike in the burned area detection importance, most features seem to have little to no influence on the prediction task. Detecting the current burned area with bands I2 and M11 is a prerequisite to predicting where the next day's burned area will be, so it makes sense that these two features have a high importance. However, the model seems to fail to make use of the other features to be able to predict how exactly the fire will spread. There are three potential reasons for this low contribution of auxiliary data. Firstly, the temporal and spatial resolution of the data may be too low to make these predictions. Secondly, the segmentation models may not be able to extract useful features from the auxiliary data for prediction. Finally, naive concatenation of auxiliary data and the spectral bands may not be the best option for data fusion.

\subsection*{Qualitative results}
\subsubsection*{Active Fire Detection}

\begin{figure*}[!hbt]
    \centering
    \includegraphics[scale=0.45]{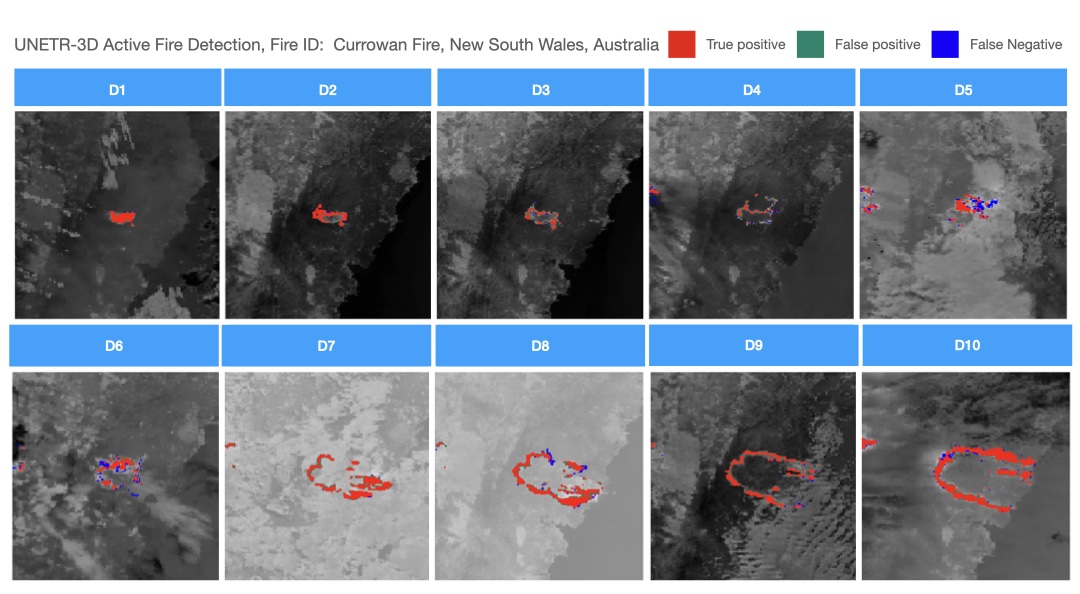}
    \caption{Results of UNETR-3D for active fire detection task. Fire ID: Currowan Fire, New South Wales, Australia.}
    \label{fig:AF2}
\end{figure*}

The qualitative results are shown in Figure \ref{fig:AF2}, the classified active fire are overlayed on top of the Band I4 images. For Figure \ref{fig:AF2}, false negative detection is the major limitation of the used model.
\subsubsection*{Burned Area Mapping}

\begin{figure*}[!hbt]
    \centering
    \includegraphics[scale=0.45]{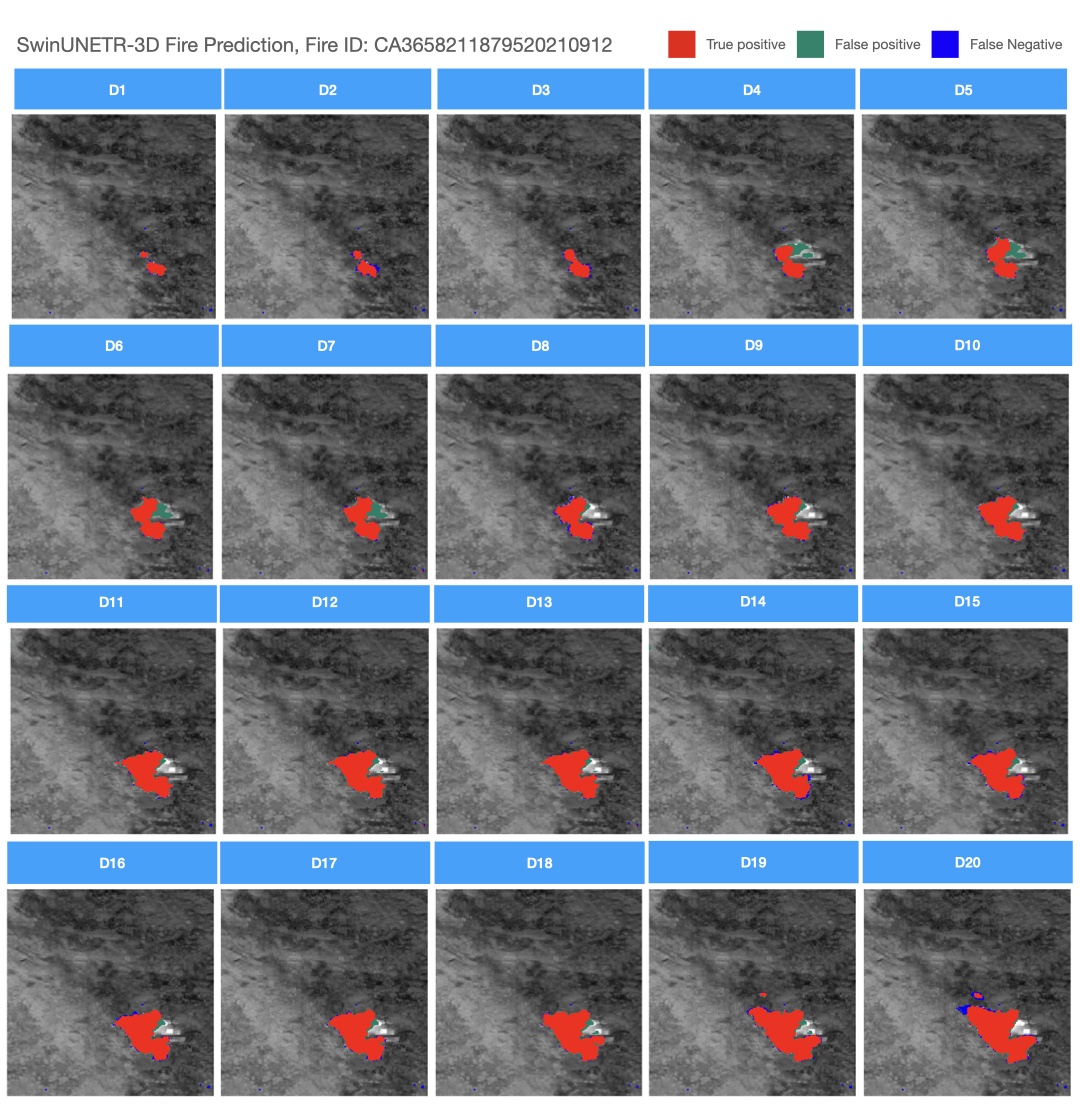}
    \caption{Results of SwinUNETR for burned area mapping task. Fire ID: CA3658211879520210912.}
    \label{fig:BA3}
\end{figure*}
For the burned area mapping task, the qualitative results are shown in Figure \ref{fig:BA3}. The classification map is overlaid on top of accumulated Band I4 images. For the three study regions presented, all burned areas are well detected.
\subsubsection*{Fire prediction}

\begin{figure*}[!hbt]
    \centering
    \includegraphics[scale=0.45]{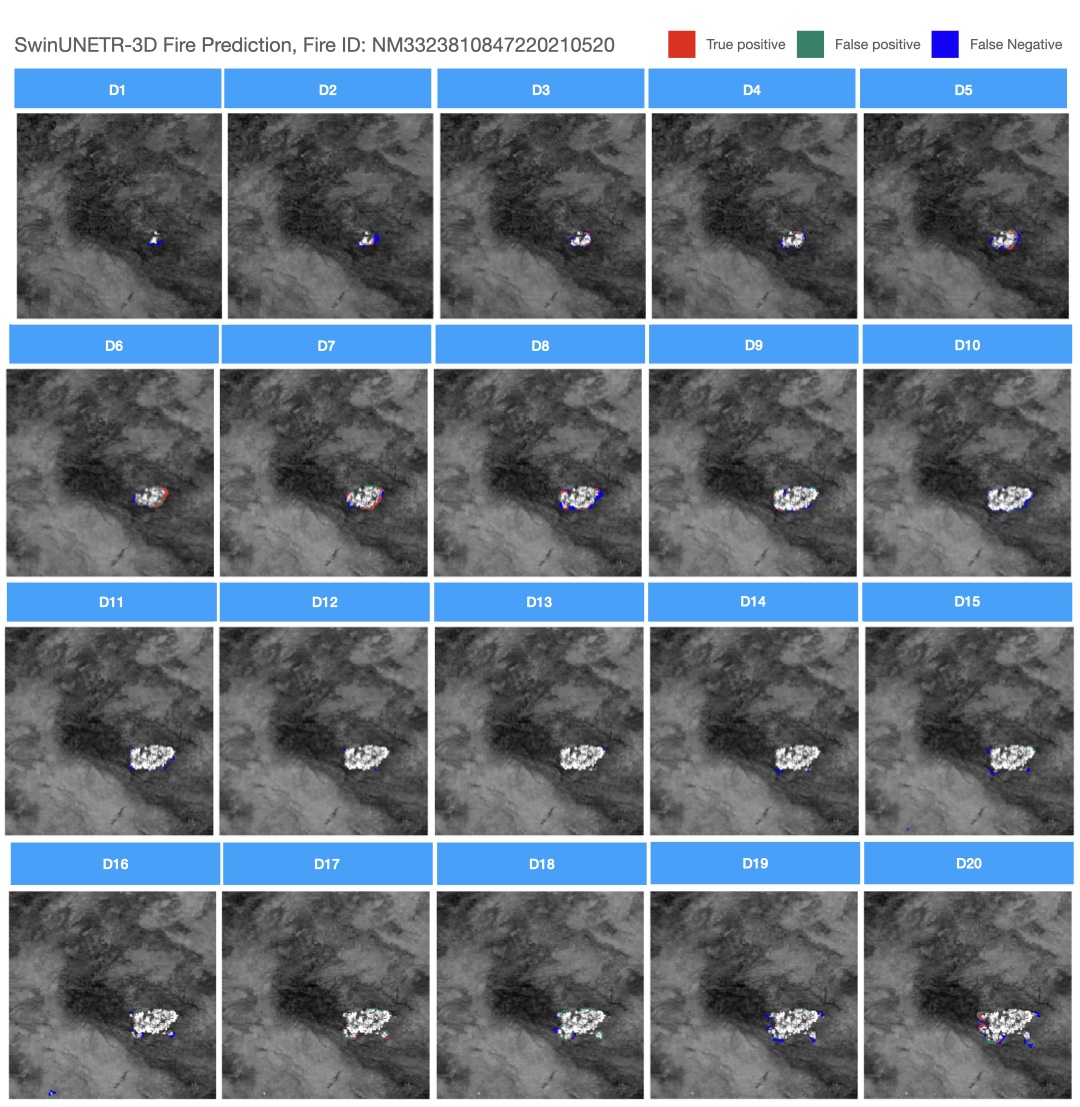}
    \caption{Results of SwinUNETR for fire progression prediction task. Fire ID: NM3323810847220210520.}
    \label{fig:pred2}
\end{figure*}

For the fire prediction task, the same study regions as the burned area mapping task are used. From Figure \ref{fig:pred2}, it can be observed that the results from SwinUNETR contain many false positive and false negative predictions. Note that the prediction model only predicts the newly burned area of each day, leading to the white areas in the center of the fire not being part of the daily prediction. Although the model makes some predictions around the burned area boundaries, the accuracy is limited. It also suggests that the prediction task is much more challenging than detection.

\section*{Usage Notes}
Example Python scripts to process GeoTIFF files into Numpy arrays for deep learning models are provided in (\textit{dataset\_gen\_afba.py} and \textit{dataset\_gen\_pred.py}) within the repository. All wildfire events and their associated information, including start and end dates as well as the region of interest, are organized in CSV files under the \textit{roi} folder, with separate files for each year.
\section*{Code Availability}
The python scripts used to process the GeoTIFF to the format consumed by the models and all the baseline models used as the benchmark are provided in GitHub Repository \url{https://github.com/zhaoyutim/TS-SatFire}.
\bibliography{sample}

\begin{thebibliography}{10}
\urlstyle{rm}
\expandafter\ifx\csname url\endcsname\relax
  \def\url#1{\texttt{#1}}\fi
\expandafter\ifx\csname urlprefix\endcsname\relax\def\urlprefix{URL }\fi
\expandafter\ifx\csname doiprefix\endcsname\relax\def\doiprefix{DOI: }\fi
\providecommand{\bibinfo}[2]{#2}
\providecommand{\eprint}[2][]{\url{#2}}

\bibitem{Andela2017AHD}
\bibinfo{author}{Andela, N.} \emph{et~al.}
\newblock \bibinfo{journal}{\bibinfo{title}{A human-driven decline in global burned area}}.
\newblock {\emph{\JournalTitle{Science}}} \textbf{\bibinfo{volume}{356}}, \bibinfo{pages}{1356 -- 1362} (\bibinfo{year}{2017}).

\bibitem{Curtis2018ClassifyingDO}
\bibinfo{author}{Curtis, P.~G.}, \bibinfo{author}{Slay, C.~M.}, \bibinfo{author}{Harris, N.~L.}, \bibinfo{author}{Tyukavina, A.} \& \bibinfo{author}{Hansen, M.~C.}
\newblock \bibinfo{journal}{\bibinfo{title}{Classifying drivers of global forest loss}}.
\newblock {\emph{\JournalTitle{Science}}} \textbf{\bibinfo{volume}{361}}, \bibinfo{pages}{1108 -- 1111} (\bibinfo{year}{2018}).

\bibitem{Tyukavina2022GlobalTO}
\bibinfo{author}{Tyukavina, A.} \emph{et~al.}
\newblock \bibinfo{title}{Global trends of forest loss due to fire from 2001 to 2019}.
\newblock In \emph{\bibinfo{booktitle}{Frontiers in Remote Sensing}} (\bibinfo{year}{2022}).

\bibitem{Wooster2021SatelliteRS}
\bibinfo{author}{Wooster, M.~J.} \emph{et~al.}
\newblock \bibinfo{journal}{\bibinfo{title}{Satellite remote sensing of active fires: History and current status, applications and future requirements}}.
\newblock {\emph{\JournalTitle{Remote Sensing of Environment}}}  (\bibinfo{year}{2021}).

\bibitem{Giglio2009AnAB}
\bibinfo{author}{Giglio, L.}, \bibinfo{author}{Loboda, T.~V.}, \bibinfo{author}{Roy, D.~P.}, \bibinfo{author}{Quayle, B.} \& \bibinfo{author}{Justice, C.~O.}
\newblock \bibinfo{journal}{\bibinfo{title}{An active-fire based burned area mapping algorithm for the {MODIS} sensor}}.
\newblock {\emph{\JournalTitle{Remote Sensing of Environment}}} \textbf{\bibinfo{volume}{113}}, \bibinfo{pages}{408--420} (\bibinfo{year}{2009}).

\bibitem{viirs}
\bibinfo{author}{Schroeder, W.}, \bibinfo{author}{Oliva, P.}, \bibinfo{author}{Giglio, L.} \& \bibinfo{author}{Csiszar, I.}
\newblock \bibinfo{journal}{\bibinfo{title}{The new {VIIRS} 375 m active fire detection data product: Algorithm description and initial assessment}}.
\newblock {\emph{\JournalTitle{Remote Sensing of Environment}}}  (\bibinfo{year}{2014}).

\bibitem{modis}
\bibinfo{author}{Schroeder, W.} \emph{et~al.}
\newblock \bibinfo{journal}{\bibinfo{title}{Validation of {GOES} and {MODIS} active fire detection products using {ASTER} and {ETM+} data}}.
\newblock {\emph{\JournalTitle{Remote Sensing of Environment}}}  (\bibinfo{year}{2008}).

\bibitem{col6}
\bibinfo{author}{Giglio, L.}, \bibinfo{author}{Schroeder, W.} \& \bibinfo{author}{Justice, C.}
\newblock \bibinfo{journal}{\bibinfo{title}{The collection 6 {MODIS} active fire detection algorithm and fire products}}.
\newblock {\emph{\JournalTitle{Remote sensing of environment}}}  (\bibinfo{year}{2016}).

\bibitem{LizundiaLoiola2020ASA}
\bibinfo{author}{Lizundia-Loiola, J.}, \bibinfo{author}{Ot{\'o}n, G.}, \bibinfo{author}{Ramo, R.} \& \bibinfo{author}{Chuvieco, E.}
\newblock \bibinfo{journal}{\bibinfo{title}{A spatio-temporal active-fire clustering approach for global burned area mapping at 250 m from modis data}}.
\newblock {\emph{\JournalTitle{Remote Sensing of Environment}}} \textbf{\bibinfo{volume}{236}}, \bibinfo{pages}{111493} (\bibinfo{year}{2020}).

\bibitem{de2021active}
\bibinfo{author}{de~Almeida~Pereira, G.~H.}, \bibinfo{author}{Fusioka, A.~M.}, \bibinfo{author}{Nassu, B.~T.} \& \bibinfo{author}{Minetto, R.}
\newblock \bibinfo{journal}{\bibinfo{title}{Active fire detection in {Landsat-8} imagery: A large-scale dataset and a deep-learning study}}.
\newblock {\emph{\JournalTitle{ISPRS Journal of Photogrammetry and Remote Sensing}}} \textbf{\bibinfo{volume}{178}}, \bibinfo{pages}{171--186} (\bibinfo{year}{2021}).

\bibitem{Ban2020NearRW}
\bibinfo{author}{Ban, Y.}, \bibinfo{author}{Zhang, P.}, \bibinfo{author}{Nascetti, A.}, \bibinfo{author}{Bevington, A.~R.} \& \bibinfo{author}{Wulder, M.}
\newblock \bibinfo{journal}{\bibinfo{title}{Near real-time wildfire progression monitoring with {Sentinel-1} {SAR} time series and deep learning}}.
\newblock {\emph{\JournalTitle{Scientific Reports}}} \textbf{\bibinfo{volume}{10}} (\bibinfo{year}{2020}).

\bibitem{NEURIPS2023_ebd54517}
\bibinfo{author}{Gerard, S.}, \bibinfo{author}{Zhao, Y.} \& \bibinfo{author}{Sullivan, J.}
\newblock \bibinfo{title}{Wildfirespreadts: A dataset of multi-modal time series for wildfire spread prediction}.
\newblock In \bibinfo{editor}{Oh, A.} \emph{et~al.} (eds.) \emph{\bibinfo{booktitle}{Advances in Neural Information Processing Systems}}, vol.~\bibinfo{volume}{36}, \bibinfo{pages}{74515--74529} (\bibinfo{publisher}{Curran Associates, Inc.}, \bibinfo{year}{2023}).

\bibitem{9840400}
\bibinfo{author}{Huot, F.} \emph{et~al.}
\newblock \bibinfo{journal}{\bibinfo{title}{Next day wildfire spread: A machine learning dataset to predict wildfire spreading from remote-sensing data}}.
\newblock {\emph{\JournalTitle{IEEE Transactions on Geoscience and Remote Sensing}}} \textbf{\bibinfo{volume}{60}}, \bibinfo{pages}{1--13}, \doiprefix\url{10.1109/TGRS.2022.3192974} (\bibinfo{year}{2022}).

\bibitem{Justice2002TheMF}
\bibinfo{author}{Justice, C.~O.} \emph{et~al.}
\newblock \bibinfo{journal}{\bibinfo{title}{The modis fire products}}.
\newblock {\emph{\JournalTitle{Remote Sensing of Environment}}} \textbf{\bibinfo{volume}{83}}, \bibinfo{pages}{244--262} (\bibinfo{year}{2002}).

\bibitem{Giglio2003AnEC}
\bibinfo{author}{Giglio, L.}, \bibinfo{author}{Descloitres, J.}, \bibinfo{author}{Justice, C.~O.} \& \bibinfo{author}{Kaufman, Y.~J.}
\newblock \bibinfo{journal}{\bibinfo{title}{An enhanced contextual fire detection algorithm for {MODIS}}}.
\newblock {\emph{\JournalTitle{Remote Sensing of Environment}}} \textbf{\bibinfo{volume}{87}}, \bibinfo{pages}{273--282} (\bibinfo{year}{2003}).

\bibitem{Xu2020FirstSO}
\bibinfo{author}{Xu, W.}, \bibinfo{author}{Wooster, M.}, \bibinfo{author}{He, J.} \& \bibinfo{author}{Zhang, T.}
\newblock \bibinfo{journal}{\bibinfo{title}{First study of {S}entinel-3 {SLSTR} active fire detection and {FRP} retrieval: Night-time algorithm enhancements and global intercomparison to {MODIS} and {VIIRS} {AF} products}}.
\newblock {\emph{\JournalTitle{Remote Sensing of Environment}}} \textbf{\bibinfo{volume}{248}}, \bibinfo{pages}{111947} (\bibinfo{year}{2020}).

\bibitem{Oliva2015AssessmentOV}
\bibinfo{author}{Oliva, P.} \& \bibinfo{author}{Schroeder, W.}
\newblock \bibinfo{journal}{\bibinfo{title}{Assessment of {VIIRS} 375 m active fire detection product for direct burned area mapping}}.
\newblock {\emph{\JournalTitle{Remote Sensing of Environment}}} \textbf{\bibinfo{volume}{160}}, \bibinfo{pages}{144--155} (\bibinfo{year}{2015}).

\bibitem{Schroeder2014TheNV}
\bibinfo{author}{Schroeder, W.}, \bibinfo{author}{Oliva, P.}, \bibinfo{author}{Giglio, L.} \& \bibinfo{author}{Csiszar, I.}
\newblock \bibinfo{journal}{\bibinfo{title}{The new {VIIRS} 375 m active fire detection data product: Algorithm description and initial assessment}}.
\newblock {\emph{\JournalTitle{Remote Sensing of Environment}}} \textbf{\bibinfo{volume}{143}}, \bibinfo{pages}{85--96} (\bibinfo{year}{2014}).

\bibitem{Zhu2017DeepLI}
\bibinfo{author}{Zhu, X.} \emph{et~al.}
\newblock \bibinfo{journal}{\bibinfo{title}{Deep learning in remote sensing: a review}}.
\newblock {\emph{\JournalTitle{ArXiv}}} \textbf{\bibinfo{volume}{abs/1710.03959}} (\bibinfo{year}{2017}).

\bibitem{Zhao2022GOESRTS}
\bibinfo{author}{Zhao, Y.} \& \bibinfo{author}{Ban, Y.}
\newblock \bibinfo{journal}{\bibinfo{title}{{GOES-R} time series for early detection of wildfires with deep {GRU}-network}}.
\newblock {\emph{\JournalTitle{Remote. Sens.}}} \textbf{\bibinfo{volume}{14}}, \bibinfo{pages}{4347} (\bibinfo{year}{2022}).

\bibitem{Zhao2023TokenizedTI}
\bibinfo{author}{Zhao, Y.}, \bibinfo{author}{Ban, Y.} \& \bibinfo{author}{Sullivan, J.}
\newblock \bibinfo{journal}{\bibinfo{title}{Tokenized time-series in satellite image segmentation with transformer network for active fire detection}}.
\newblock {\emph{\JournalTitle{IEEE Transactions on Geoscience and Remote Sensing}}} \textbf{\bibinfo{volume}{61}}, \bibinfo{pages}{1--13} (\bibinfo{year}{2023}).

\bibitem{Roy2002BurnedAM}
\bibinfo{author}{Roy, D.~P.}, \bibinfo{author}{Lewis, P.} \& \bibinfo{author}{Justice, C.~O.}
\newblock \bibinfo{journal}{\bibinfo{title}{Burned area mapping using multi-temporal moderate spatial resolution data—a bi-directional reflectance model-based expectation approach}}.
\newblock {\emph{\JournalTitle{Remote Sensing of Environment}}} \textbf{\bibinfo{volume}{83}}, \bibinfo{pages}{263--286} (\bibinfo{year}{2002}).

\bibitem{Roy1999MultitemporalAB}
\bibinfo{author}{Roy, D.~P.}
\newblock \bibinfo{journal}{\bibinfo{title}{Multi-temporal active-fire based burn scar detection algorithm}}.
\newblock {\emph{\JournalTitle{International Journal of Remote Sensing}}} \textbf{\bibinfo{volume}{20}}, \bibinfo{pages}{1031--1038} (\bibinfo{year}{1999}).

\bibitem{Scaduto2020SatelliteBasedFP}
\bibinfo{author}{Scaduto, E.}, \bibinfo{author}{Chen, B.} \& \bibinfo{author}{Jin, Y.}
\newblock \bibinfo{journal}{\bibinfo{title}{Satellite-based fire progression mapping: A comprehensive assessment for large fires in northern california}}.
\newblock {\emph{\JournalTitle{IEEE Journal of Selected Topics in Applied Earth Observations and Remote Sensing}}} \textbf{\bibinfo{volume}{13}}, \bibinfo{pages}{5102--5114} (\bibinfo{year}{2020}).

\bibitem{Chuvieco2018GenerationAA}
\bibinfo{author}{Chuvieco, E.} \emph{et~al.}
\newblock \bibinfo{journal}{\bibinfo{title}{Generation and analysis of a new global burned area product based on modis 250 m reflectance bands and thermal anomalies}}.
\newblock {\emph{\JournalTitle{Earth System Science Data}}}  (\bibinfo{year}{2018}).

\bibitem{Zhang2023TotalvariationRU}
\bibinfo{author}{Zhang, P.}, \bibinfo{author}{Ban, Y.} \& \bibinfo{author}{Nascetti, A.}
\newblock \bibinfo{journal}{\bibinfo{title}{Total-variation regularized u-net for wildfire burned area mapping based on sentinel-1 c-band sar backscattering data}}.
\newblock {\emph{\JournalTitle{ISPRS Journal of Photogrammetry and Remote Sensing}}}  (\bibinfo{year}{2023}).

\bibitem{Zhao2022GlobalSB}
\bibinfo{author}{Zhao, Y.} \& \bibinfo{author}{Ban, Y.}
\newblock \bibinfo{journal}{\bibinfo{title}{Global scale burned area mapping using bi-temporal alos-2 palsar-2 l-band data}}.
\newblock {\emph{\JournalTitle{IGARSS 2022 - 2022 IEEE International Geoscience and Remote Sensing Symposium}}} \bibinfo{pages}{695--698} (\bibinfo{year}{2022}).

\bibitem{Knopp2020ADL}
\bibinfo{author}{Knopp, L.}, \bibinfo{author}{Wieland, M.}, \bibinfo{author}{R{\"a}ttich, M.} \& \bibinfo{author}{Martinis, S.}
\newblock \bibinfo{journal}{\bibinfo{title}{A deep learning approach for burned area segmentation with {Sentinel-2} data}}.
\newblock {\emph{\JournalTitle{Remote. Sens.}}} \textbf{\bibinfo{volume}{12}}, \bibinfo{pages}{2422} (\bibinfo{year}{2020}).

\bibitem{Pereira2021ActiveFD}
\bibinfo{author}{Pereira, G.}, \bibinfo{author}{Fusioka, A.~M.}, \bibinfo{author}{Nassu, B.~T.} \& \bibinfo{author}{Minetto, R.}
\newblock \bibinfo{journal}{\bibinfo{title}{Active fire detection in {Landsat-8} imagery: a large-scale dataset and a deep-learning study}}.
\newblock {\emph{\JournalTitle{ArXiv}}} \textbf{\bibinfo{volume}{abs/2101.03409}} (\bibinfo{year}{2021}).

\bibitem{Gmez2011PrototypingAA}
\bibinfo{author}{G{\'o}mez, I.} \& \bibinfo{author}{Mart{\'i}n, M.~P.}
\newblock \bibinfo{journal}{\bibinfo{title}{Prototyping an artificial neural network for burned area mapping on a regional scale in mediterranean areas using modis images}}.
\newblock {\emph{\JournalTitle{Int. J. Appl. Earth Obs. Geoinformation}}} \textbf{\bibinfo{volume}{13}}, \bibinfo{pages}{741--752} (\bibinfo{year}{2011}).

\bibitem{PINTO2020260}
\bibinfo{author}{Pinto, M.~M.}, \bibinfo{author}{Libonati, R.}, \bibinfo{author}{Trigo, R.~M.}, \bibinfo{author}{Trigo, I.~F.} \& \bibinfo{author}{DaCamara, C.~C.}
\newblock \bibinfo{journal}{\bibinfo{title}{A deep learning approach for mapping and dating burned areas using temporal sequences of satellite images}}.
\newblock {\emph{\JournalTitle{ISPRS Journal of Photogrammetry and Remote Sensing}}} \textbf{\bibinfo{volume}{160}}, \bibinfo{pages}{260--274}, \doiprefix\url{https://doi.org/10.1016/j.isprsjprs.2019.12.014} (\bibinfo{year}{2020}).

\bibitem{Kondylatos2023MesogeosAM}
\bibinfo{author}{Kondylatos, S.}, \bibinfo{author}{Prapas, I.}, \bibinfo{author}{Camps-Valls, G.} \& \bibinfo{author}{Papoutsis, I.}
\newblock \bibinfo{journal}{\bibinfo{title}{Mesogeos: A multi-purpose dataset for data-driven wildfire modeling in the mediterranean}}.
\newblock {\emph{\JournalTitle{ArXiv}}} \textbf{\bibinfo{volume}{abs/2306.05144}} (\bibinfo{year}{2023}).

\bibitem{finney_farsite_1998}
\bibinfo{author}{Finney, M.~A.}
\newblock \bibinfo{journal}{\bibinfo{title}{{FARSITE}: Fire area simulator-model development and evaluation}}.
\newblock {\emph{\JournalTitle{Res. Pap. {RMRS}-{RP}-4, Revised 2004. Ogden, {UT}: U.S. Department of Agriculture, Forest Service, Rocky Mountain Research Station. 47 p.}}} \textbf{\bibinfo{volume}{4}}, \doiprefix\url{10.2737/RMRS-RP-4} (\bibinfo{year}{1998}).

\bibitem{tymstra_development_2010}
\bibinfo{author}{Tymstra, C.}, \bibinfo{author}{Bryce, R.~W.}, \bibinfo{author}{Wotton, B.~M.}, \bibinfo{author}{Taylor, S.~W.} \& \bibinfo{author}{Armitage, O.~B.}
\newblock \emph{\bibinfo{title}{Development and structure of Prometheus: the Canadian Wildland Fire Growth Simulation Model.}}, vol. \bibinfo{volume}{417} (\bibinfo{publisher}{Natural Resources Canada}, \bibinfo{year}{2010}).
\newblock \bibinfo{note}{{ISSN}: 0831-8247}.

\bibitem{Radke2019FireCastLD}
\bibinfo{author}{Radke, D.}, \bibinfo{author}{Hessler, A.} \& \bibinfo{author}{Ellsworth, D.}
\newblock \bibinfo{title}{Firecast: Leveraging deep learning to predict wildfire spread}.
\newblock In \emph{\bibinfo{booktitle}{International Joint Conference on Artificial Intelligence}} (\bibinfo{year}{2019}).

\bibitem{singla_wildfiredb_2021}
\bibinfo{author}{Singla, S.} \emph{et~al.}
\newblock \bibinfo{title}{{WildfireDB}: An open-source dataset connecting wildfire occurrence with relevant determinants}.
\newblock In \emph{\bibinfo{booktitle}{Thirty-fifth Conference on Neural Information Processing Systems Datasets and Benchmarks Track (Round 2)}} (\bibinfo{year}{2021}).

\bibitem{huot_next_2022}
\bibinfo{author}{Huot, F.} \emph{et~al.}
\newblock \bibinfo{journal}{\bibinfo{title}{Next day wildfire spread: A machine learning dataset to predict wildfire spreading from remote-sensing data}}.
\newblock {\emph{\JournalTitle{{IEEE} Transactions on Geoscience and Remote Sensing}}} \textbf{\bibinfo{volume}{60}}, \bibinfo{pages}{1--13}, \doiprefix\url{10.1109/TGRS.2022.3192974} (\bibinfo{year}{2022}).
\newblock \bibinfo{note}{Conference Name: {IEEE} Transactions on Geoscience and Remote Sensing}.

\bibitem{Gerard2023WildfireSpreadTSAD}
\bibinfo{author}{Gerard, S.}, \bibinfo{author}{Zhao, Y.} \& \bibinfo{author}{Sullivan, J.}
\newblock \bibinfo{title}{{WildfireSpreadTS}: A dataset of multi-modal time series for wildfire spread prediction}.
\newblock In \emph{\bibinfo{booktitle}{Neural Information Processing Systems}} (\bibinfo{year}{2023}).

\bibitem{gee}
\bibinfo{author}{Gorelick, N.} \emph{et~al.}
\newblock \bibinfo{journal}{\bibinfo{title}{Google earth engine: Planetary-scale geospatial analysis for everyone}}.
\newblock {\emph{\JournalTitle{Remote Sensing of Environment}}}  (\bibinfo{year}{2017}).

\bibitem{abatzoglou_development_2013}
\bibinfo{author}{Abatzoglou, J.~T.}
\newblock \bibinfo{journal}{\bibinfo{title}{Development of gridded surface meteorological data for ecological applications and modelling}}.
\newblock {\emph{\JournalTitle{International Journal of Climatology}}} \textbf{\bibinfo{volume}{33}} (\bibinfo{year}{2013}).

\bibitem{clough_atmospheric_2005}
\bibinfo{author}{Clough, S.~A.} \emph{et~al.}
\newblock \bibinfo{journal}{\bibinfo{title}{Atmospheric radiative transfer modeling: a summary of the {AER} codes}}.
\newblock {\emph{\JournalTitle{Journal of Quantitative Spectroscopy and Radiative Transfer}}} \textbf{\bibinfo{volume}{91}}, \bibinfo{pages}{233--244}, \doiprefix\url{10.1016/j.jqsrt.2004.05.058} (\bibinfo{year}{2005}).

\bibitem{Littell2016ARO}
\bibinfo{author}{Littell, J.~S.}, \bibinfo{author}{Peterson, D.~L.}, \bibinfo{author}{Riley, K.~L.}, \bibinfo{author}{Liu, Y.} \& \bibinfo{author}{Luce, C.~H.}
\newblock \bibinfo{journal}{\bibinfo{title}{A review of the relationships between drought and forest fire in the united states}}.
\newblock {\emph{\JournalTitle{Global Change Biology}}} \textbf{\bibinfo{volume}{22}} (\bibinfo{year}{2016}).

\bibitem{Brown2023ClimateWI}
\bibinfo{author}{Brown, P.~T.} \emph{et~al.}
\newblock \bibinfo{journal}{\bibinfo{title}{Climate warming increases extreme daily wildfire growth risk in california}}.
\newblock {\emph{\JournalTitle{Nature}}} \textbf{\bibinfo{volume}{621}}, \bibinfo{pages}{760--766} (\bibinfo{year}{2023}).

\bibitem{Povak2018EvidenceFS}
\bibinfo{author}{Povak, N.~A.}, \bibinfo{author}{Hessburg, P.~F.} \& \bibinfo{author}{Salter, R.~B.}
\newblock \bibinfo{journal}{\bibinfo{title}{Evidence for scale‐dependent topographic controls on wildfire spread}}.
\newblock {\emph{\JournalTitle{Ecosphere}}}  (\bibinfo{year}{2018}).

\bibitem{nasa_jpl_nasa_2013}
\bibinfo{author}{{JPL}, N.}
\newblock \bibinfo{journal}{\bibinfo{title}{{NASA} shuttle radar topography mission global 1 arc second}}.
\newblock {\emph{\JournalTitle{{NASA} {EOSDIS} Land Processes {DAAC}}}} \doiprefix\url{10.5067/MEaSUREs/SRTM/SRTMGL1.003} (\bibinfo{year}{2013}).

\bibitem{sulla-menashe_hierarchical_2019}
\bibinfo{author}{Sulla-Menashe, D.}, \bibinfo{author}{Gray, J.~M.}, \bibinfo{author}{Abercrombie, S.~P.} \& \bibinfo{author}{Friedl, M.~A.}
\newblock \bibinfo{journal}{\bibinfo{title}{Hierarchical mapping of annual global land cover 2001 to present: The {MODIS} collection 6 land cover product}}.
\newblock {\emph{\JournalTitle{Remote Sensing of Environment}}} \textbf{\bibinfo{volume}{222}}, \bibinfo{pages}{183--194} (\bibinfo{year}{2019}).

\bibitem{friedl_modisterraaqua_2022}
\bibinfo{author}{Friedl, M.~A.} \& \bibinfo{author}{Sulla-Menashe, D.}
\newblock \bibinfo{journal}{\bibinfo{title}{{MODIS}/terra+aqua land cover type yearly l3 global 500m {SIN} grid v061}}.
\newblock {\emph{\JournalTitle{{NASA} {EOSDIS} Land Processes {DAAC}}}} \doiprefix\url{10.5067/MODIS/MCD12Q1.061} (\bibinfo{year}{2022}).

\bibitem{yu_zhao_2024}
\bibinfo{author}{Zhao, Y.}
\newblock \bibinfo{title}{Ts-satfire}, \doiprefix\url{10.34740/KAGGLE/DSV/8675553} (\bibinfo{year}{2024}).

\bibitem{unet}
\bibinfo{author}{Ronneberger, O.}, \bibinfo{author}{Fischer, P.} \& \bibinfo{author}{Brox, T.}
\newblock \bibinfo{title}{{U-Net}: Convolutional networks for biomedical image segmentation}.
\newblock In \emph{\bibinfo{booktitle}{MICCAI}} (\bibinfo{year}{2015}).

\bibitem{hatamizadeh2021unetr}
\bibinfo{author}{Hatamizadeh, A.} \emph{et~al.}
\newblock \bibinfo{title}{Unetr: Transformers for 3d medical image segmentation} (\bibinfo{year}{2021}).
\newblock \eprint{2103.10504}.

\bibitem{Tang2021SelfSupervisedPO}
\bibinfo{author}{Tang, Y.} \emph{et~al.}
\newblock \bibinfo{journal}{\bibinfo{title}{Self-supervised pre-training of swin transformers for 3d medical image analysis}}.
\newblock {\emph{\JournalTitle{2022 IEEE/CVF Conference on Computer Vision and Pattern Recognition (CVPR)}}} \bibinfo{pages}{20698--20708} (\bibinfo{year}{2021}).

\bibitem{Liu2021SwinTH}
\bibinfo{author}{Liu, Z.} \emph{et~al.}
\newblock \bibinfo{journal}{\bibinfo{title}{Swin transformer: Hierarchical vision transformer using shifted windows}}.
\newblock {\emph{\JournalTitle{2021 IEEE/CVF International Conference on Computer Vision (ICCV)}}} \bibinfo{pages}{9992--10002} (\bibinfo{year}{2021}).

\end{thebibliography}
\section*{Acknowledgements}

The research is part of the project ‘Sentinel4Wildfire’ funded by Formas, the Swedish research council for sustainable development and the project ‘EO-AI4Global Change’ funded by Digital Futures.

\section*{Author contributions statement}

Y.Z., S.G., and Y.B. contributed to Conceptualization, methodology, and validation. Y.Z. and S.G. contributed to the implementation, formal analysis, investigation, and original draft preparation. Y.B. contributed to resources, writing—review and editing, supervision, project administration, and funding acquisition.

\section*{Competing Interests}
The authors declare no competing interests.

\section*{Additional Information}
Correspondence and requests for materials should be addressed to Y.B.
\end{document}